\documentclass{article}

    \PassOptionsToPackage{numbers, compress}{natbib}

\usepackage[final]{neurips_2022}

\usepackage[utf8]{inputenc} 
\usepackage[T1]{fontenc}    
\usepackage{hyperref}       
\usepackage{url}            
\usepackage{booktabs}       
\usepackage{amsfonts}       
\usepackage{nicefrac}       
\usepackage{microtype}      
\usepackage{xcolor}         
\usepackage{subcaption} 
\usepackage{graphicx}
\usepackage{amsmath}
\usepackage{amssymb}
\usepackage{makecell}
\usepackage{multirow}
\usepackage{xcolor}
\usepackage{pgfplots}
\usepackage{pifont}
\newcommand{\cmark}{\text{\ding{51}}}
\newcommand{\xmark}{\text{\ding{55}}}
\pgfplotsset{compat=1.9}
\usepgfplotslibrary{external}
\title{Could Giant Pretrained Image Models Extract Universal Representations?}

\author{%
  Yutong Lin\thanks{Equal contribution. The work is done when Yutong Lin and Ze Liu are interns at Microsoft Research Asia. Correspondence to: Yue Cao (caoyue10@gmail.com).}\hspace{0.4mm} $^{13}$, Ze Liu$^{*23}$, Zheng Zhang$^{3}$, Han Hu$^{3}$, Nanning Zheng$^{1}$, Stephen Lin$^{3}$, Yue Cao$^{3}$\\
  $^1$Xi'an Jiaotong University\\
  $^2$University of Science and Technology of China\\
  $^3$Microsoft Research Asia\\
  \texttt{\{t-yutonglin,t-liuze,zhez,hanhu,stevelin\}@microsoft.com} \\
  \texttt{nnzheng@mail.xjtu.edu.cn,caoyue10@gmail.com}
}

\begin{document}

\maketitle

\begin{abstract}
Frozen pretrained models have become a viable alternative to the pretraining-then-finetuning paradigm for transfer learning. However, with frozen models there are relatively few parameters available for adapting to downstream tasks, which is problematic in computer vision where tasks vary significantly in input/output format and the type of information that is of value. In this paper, we present a study of frozen pretrained models when applied to diverse and representative computer vision tasks, including object detection, semantic segmentation and video action recognition. From this empirical analysis, our work answers the questions of what pretraining task fits best with this frozen setting, how to make the frozen setting more flexible to various downstream tasks, and the effect of larger model sizes. We additionally examine the upper bound of performance using a giant frozen pretrained model with 3 billion parameters (SwinV2-G) and find that it reaches competitive performance on a varied set of major benchmarks with only one shared frozen base network: 60.0 box mAP and 52.2 mask mAP on COCO object detection test-dev, 57.6 val mIoU on ADE20K semantic segmentation, and 81.7 top-1 accuracy on Kinetics-400 action recognition. With this work, we hope to bring greater attention to this promising path of freezing pretrained image models.
\end{abstract}

\section{Introduction}
Transfer learning via the pretraining-then-finetuning paradigm is the cornerstone in the success of deep neural networks. In computer vision, finetuning backbone networks~\cite{alexnet,he2016resnet,dosovitskiy2020vit,liu2021swin} pretrained on supervised classification leads to top performance for a wide range of visual recognition tasks, even on tasks where the input or output format differs from that of pretraining, such as video action recognition~\cite{kay2017kinetics}, object detection~\cite{lin2014coco}, and semantic segmentation~\cite{zhou2018ade}.
In natural language processing (NLP), finetuned models that are pretrained on masked language modeling (MLM) also excel at a large variety of NLP tasks, e.g., on eleven tasks in the case of BERT~\cite{devlin2018bert}.

Though highly effective, the changes to many network parameters make finetuning parameter-inefficient. An entirely different model is created for each downstream task, a problem that is magnified with the rapid increase in model sizes. In NLP, a solution to this problem that has gained momentum is to freeze pretrained language models when transferring them to downstream tasks. A small number of task-specific parameters, e.g., task-specific heads~\cite{houlsby2019parameter} or prompts~\cite{gpt3,lester2021power,li2021prefix,liu2021ptuning}, are introduced, and only these newly added parameters are trained on the downstream task while the rest of the model is fixed. In this way, the computation and memory costs are reduced considerably, with a high degree of parameter sharing among models finetuned for different tasks.

This parameter-efficient transfer, however, has yet to gain traction in computer vision. The difficulties in adopting this approach lie in the differences between the two modalities. In NLP, the input and output formats of different tasks are similar, where nearly all of them can be defined as a sequence of tokens. Moreover, the dominant pretraining task of self-supervised masked language modeling learns information that is broadly valuable in NLP applications. By contrast, such uniformity does not exist in computer vision. The input and output formats in vision vary greatly, such as image/video and high/low resolution for inputs, and image-level categories, target coordinates and so on for outputs. The information that is valuable varies as well among downstream tasks, making the choice of pretraining task unclear. Due to these challenges, previous investigations have focused only on downstream tasks similar to the pretraining task, image classification~\cite{bilen2017universal,rebuffi2017resadapt,rebuffi2018multidomain,zhang2020side,yao2021cpt,zhou2021vprompt}.

In this paper, we study parameter-efficient transfer learning that generalizes to a diverse set of computer vision tasks. Under this setting, our work seeks to answer the following questions:
\begin{enumerate}
    \item Which pretraining task fits best with this frozen setting?
    \item What is the key to making the frozen setting work well when the downstream tasks are significantly different from the pretraining task?
    \item How well can a frozen setting perform with a giant pretrained model, such as SwinV2-G with 3B parameters?
\end{enumerate}

\begin{figure}
    \centering
    \includegraphics[width=\linewidth]{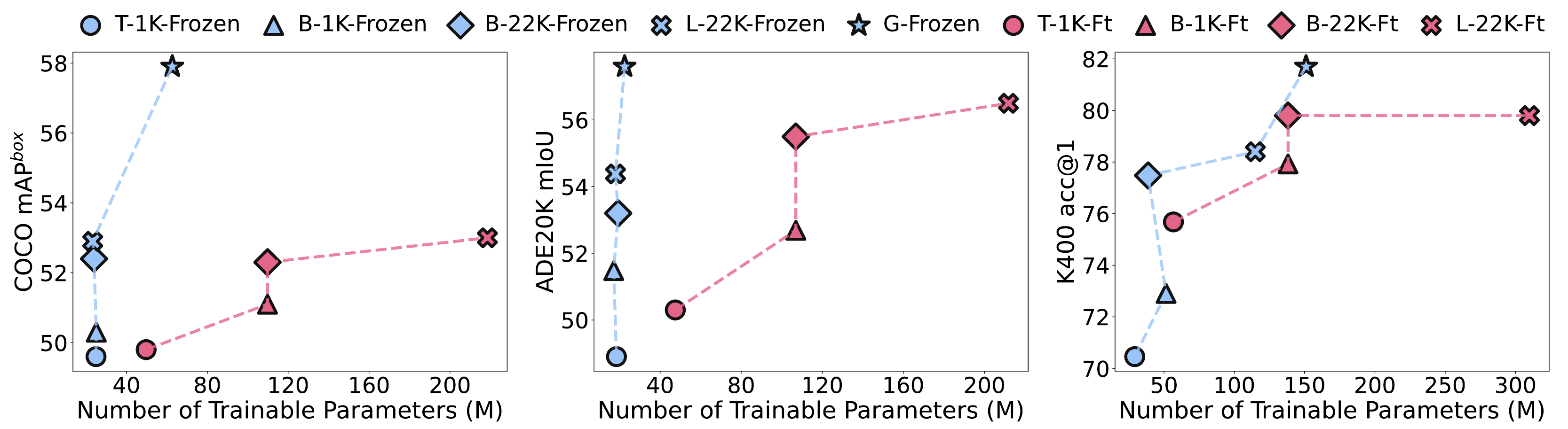}
    \caption{Performance with respect to the number of trainable parameters with different base networks (SwinV2-T/B with supervised training on ImageNet-1K, SwinV2-B/L with supervised training on ImageNet-22k, and SwinV2-G) under frozen and full finetuning settings on COCO object detection (left), ADE20K semantic segmentation (middle), and Kinetics-400 video action recognition (right). Single-scale box mAP for COCO, multi-scale mIoU for ADE20K and top-1 accuracy for Kinetics-400 are reported.}
    \label{fig:teaser}
    \vspace{-1em}
\end{figure}

Our investigation begins with the first question, where we consider the four most widely-studied pretraining tasks: supervised classification, contrastive learning, masked image modeling and image-text alignment. With SwinV2-B~\cite{liu2021swin,swinv2} as the backbone, the pretraining tasks are evaluated on popular vision frameworks covering a range of downstream tasks, including Mask R-CNN~\cite{Mask-rcnn} with FPN~\cite{FPN} for COCO object detection, Mask2former~\cite{mask2former} for ADE20K semantic segmentation, and Video Swin Transformer~\cite{liu2021video} for Kinetics-400 (K400) video action recognition. We find that supervised pretraining and image-text alignment work best under the frozen setting and the model with ImageNet-22K pretraining performs better than the ImageNet-1K counterpart. However, the frozen settings still underperform full finetuning by substantial margins.

Towards improving performance over disparate tasks, we next consider how to extend the frozen setting for better task adaptation. We examine the effects of adding tunable parameters to task-specific heads and task-based architectural elements such as FPN for object detection and decoders for semantic segmentation. From a performance analysis and an examination of feature similarity across different layers by Centered Kernel Alignment (CKA)~\cite{cka}, we make a number of observations for different tasks on what can help to bridge the performance gap between frozen settings and full finetuning. Overall, we find that the number of well-placed tunable parameters is the key to make frozen settings work well. 

Finally, we look at the impact of model size on the frozen setting. With its richer content, larger pretrained models have been found in NLP to require fewer tunable layers. Despite a large difference between pretraining and finetuning tasks as well as a relatively small change in model size, we have observed similar trends for computer vision between SwinV2-L-22K and SwinV2-T-1K in object detection. We additionally explore the upper bound of performance under the frozen setting, using a frozen supervised pretrained model with 3 billion parameters trained on 70M images (SwinV2-G). With this giant model, highly competitive performance is achieved on major benchmarks: 60.0 box AP on COCO object detection test-dev, 57.6 mIoU on ADE20K semantic segmentation, and 81.7 top-1 accuracy on Kinetics-400 action recognition. 

Thanks to the efficiency of this frozen setting, competitive performance are achieved on different model sizes with much less trainable parameters, as indicated in Figure~\ref{fig:teaser}. We hope our work will inspire further research in this promising direction of freezing pretrained image models. The proposed approach can serve as a simple baseline and guide the evaluation of future work.

\section{Related Work}

\paragraph{Representative Visual Pretraining Tasks}  Throughout the deep learning era, supervised learning, especially by image classification on ImageNet~\cite{deng2009imagenet}, has been prevalent in vision pretraining. Leveraging a large amount of data, a supervised pretrained model can efficiently transfer to various downstream tasks~\cite{donahue2014decaf,kornblith2019better,alex2019big,dosovitskiy2020vit,sermanet2013overfeat,girshick2014rich,liu2021swin,long2015fully,simonyan2014two,carreira2017quo} by finetuning. However, supervised pretraining suffers from severe data-hunger, while the cost of acquiring labeled data is expensive. To address this issue, several recent studies in self-supervised pretraining have demonstrated promising results, where self-supervised pretraining achieves finetuning performance on par with the supervised counterparts on several representative downstream tasks~\cite{he2019moco,chen2020simclr}. Among them, contrastive learning~\cite{dosovitskiy2014exemplarcnn,he2019moco,chen2020simclr,cao2020pic,grill2020byol} and masked image modeling~\cite{chen2020imagegpt,bao2021beit,he2021masked,xie2021simmim} have been particularly successful. Contrastive learning compares two image views, maximizing the similarity of positive pairs while minimizing the similarity of negative pairs. For linear probing, the state-of-the-art contrastive model even achieves results comparable to supervised models. Differently, masked image modeling learns by randomly masking some input tokens and reconstructing the raw signals. With masked image modeling, a large-scale model can be trained more effectively and yields better finetuning performance. Another branches of work utilize image-text pairs and leverage natural language as supervision for visual-linguistic learning. Among them, image-text alignment pre-training~\cite{radford2021clip, jia2021align} with web-crawled data have shown great potential for visual recognition tasks, especially under the zero-shot setting. iCar~\cite{yixuan2022icar} further bridges image-text alignment and image classification and successfully keep the characters of both methods.

These pretraining methods are so successful that almost all the top models of various vision tasks are finetuned from them. However, in the frozen backbone setting, the performance of these pretraining tasks for various downstream tasks are still unknown. 
Based on this motivation, we carried out a comparison study under the frozen setting with the four most widely-used pretraining tasks -- supervised classification~\cite{liu2021swin,swinv2}, contrastive learning~\cite{esvit},  masked image modeling~\cite{xie2021simmim} and image-text alignment~\cite{yixuan2022icar} -- using Swin Transformer as the backbone.

\paragraph{Frozen Language Models} With the large scale of today's language models, the setting of frozen language models has become important in NLP and receives much attention~\cite{liu2021pre}. Adding external adapters~\cite{houlsby2019parameter} is a direct solution in this direction and was first introduced in this field. Subsequently, various solutions specific to NLP have been proposed, including prompt tuning~\cite{lester2021power}, prefix tuning~\cite{li2021prefix}, and low-rank adaptation~\cite{hu2021lora}.

\paragraph{Frozen Setting in Vision} In computer vision, additive models are the most studied direction for the frozen setting, where the pretrained weights are frozen and a small number of new parameters is added for each task. Some of these works~\cite{donahue2014decaf,sharif2014cnn} directly add the new layers upon off-the-shelf features, while others introduce a new network with independent access to the input~\cite{rusu2016progressive,zhang2020side} to address the information loss of the pretrained model, or add task-specific components into the pretrained model~\cite{rebuffi2017resadapt,bilen2017universal,rebuffi2018multidomain}, such as batch normalization, residual adapter and so on. However, all previous works~\cite{bilen2017universal,rebuffi2017resadapt,rebuffi2018multidomain,zhang2020side} use the additive models only for tasks similar to the image classification pretraining task. In concurrent work~\cite{vasconcelos2022proper}, image classification features are reused only for object detection. Our paper, for the first time, studies the frozen setting under different pretrained image models on three representative but diverse vision tasks, namely object detection, semantic segmentation and video action recognition. Furthermore, we demonstrate the effectiveness of our approach via a giant frozen supervised pretrained model (SwinV2-G), with highly competitive performance on major benchmarks for the three tasks.
Another path is to directly adapt the prompt learning approach in NLP to computer vision~\cite{zhou2021vprompt,yao2021cpt}, but this is limited to very specific tasks or scenarios, e.g., requiring a text encoder~\cite{zhou2021vprompt}.

\section{Methodology}
Under the frozen setting, we take the approach of freezing the pretrained base network B($\cdot$) and training a specially designed task-specific head network H($\cdot$) for each downstream task.

\subsection{Architectural Elements}

\paragraph{Base Network} We denote deep neural networks designed for image classification, e.g., VGG~\cite{simonyan2014vgg}, ResNet~\cite{he2016resnet} or ViT~\cite{dosovitskiy2020vit}, as the \emph{base network}, B($\cdot$). The base network is also referred to as the backbone network in some scenarios, and it is intended to provide some core cognitive or perceptual information about the input. In the frozen setting, after the base network B($\cdot$) is pretrained, we do not update it when it is transferred to downstream tasks, that is, the base network has no tunable parameters for training on downstream tasks.

For the base network, we adopt a general-purpose backbone architecture, e.g., Swin Transformer~\cite{liu2021swin} for most experiments, that is compatible with a broad range of vision tasks, without components specific to any downstream tasks. We note that Swin Transformer is highly scalable, with a giant version composed of 3 billion parameters~\cite{swinv2}.

\paragraph{Head Network} We denote the task-specific network that is newly added and tuned during transfer to downstream tasks as the \emph{head network}, H($\cdot$). Note that the input and output formats in vision vary greatly, so a crucial component of our approach is to effectively adapt the head network to the properties of the downstream task.
With a suitable head design, we find this simple framework to generally perform well over the sampled downstream tasks under the frozen pretrained image model setting.

\begin{figure}
    \centering
    \includegraphics[width=.76\linewidth]{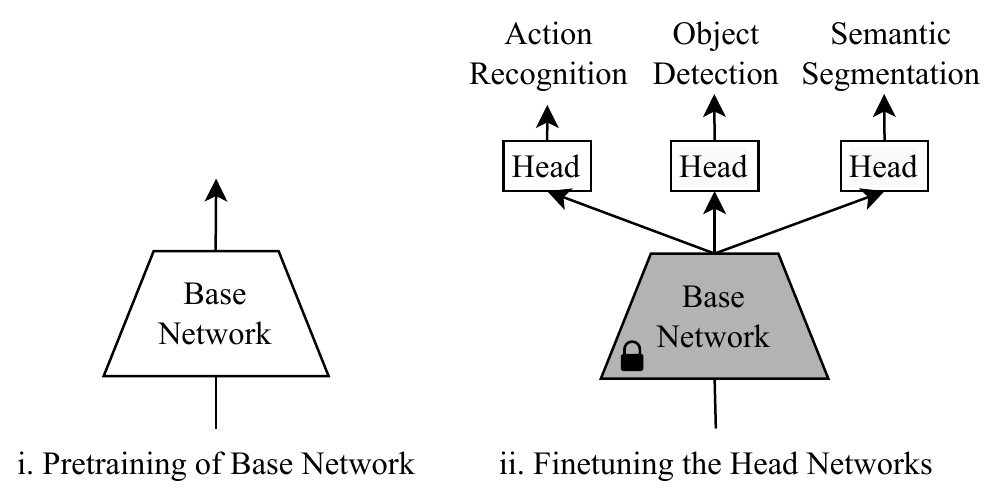}
    \caption{Pipeline of the frozen setting, consisting of (i) pretraining of the base network, and (ii) finetuning of head networks with the frozen base network.}
    \label{fig:arch}
\end{figure}

\subsection{Pretraining}

We consider the most widely-used pretraining tasks -- supervised classification~\cite{liu2021swin,swinv2}, contrastive learning, masked image modeling and visual-linguistic learning -- with Swin Transformer as the base network. Specifically, we utilize Swin Transformer variants specially designed for each of these pretraining tasks, e.g., vanilla Swin Transformer~\cite{liu2021swin,swinv2} for supervised classification, EsViT~\cite{esvit} for contrastive learning, SimMIM~\cite{xie2021simmim} for masked image modeling and iCar\cite{yixuan2022icar} for visual-linguistic learning, which are all open-source.
For fair comparison, the same base network capacity is maintained across the pretraining tasks, such as using Swin Transformer with 224$\times$224 input and a window size of 7.

\subsection{Frozen Setting on Downstream Tasks} 
A major challenge of the frozen setting is the varied input and output formats of different vision tasks, such as object detection with high-resolution images as input and target coordinates and the corresponding categories as output, semantic segmentation with high-resolution images for input and pixel-level categories as output, and video action recognition with low-resolution videos for input and video-level categories as output. 
Therefore, adapting a frozen base network to a new task requires careful model design, including a suitable framework and head network, to bridge the gap between pretraining and finetuning.

\paragraph{Object Detection} To adapt a frozen pretrained model to the object detection task, we present a few adjustments. First, object detection generally requires high-resolution images as input and a large window size in the backbone architecture, while the pretraining task takes low-resolution images and a small window size. We find that directly changing the size of input images from small to large but keeping the window size unchanged works relatively well.
Second, maintaining the aspect ratio of the input image is preferable in object detection and thus by default an arbitrary input resolution is allowed for each image, with undesirable padding on the border of feature maps that change the distributions of features. 
We deal with this issue by adopting a multi-scale augmentation similar to that of image classification. It randomly resizes the original image, then randomly crops a square part of the resized image~\cite{simple_copy_paste}.
Third, object detection typically utilizes multi-resolution feature maps as the input of the head network, while most pretraining tasks only take the final output of the base network. Using Swin Transformer as the base network enables us to directly take the output features of different stages as the input to the head network, and we find that this choice empirically works well. For evaluation on this task, we adopt COCO 2017~\cite{lin2014coco}, the most widely-used benchmark for object detection and instance segmentation, which contains 118K training, 5K validation and 20K test-dev images.

\paragraph{Semantic Segmentation} Semantic segmentation aims to perform pixel-level classification on high resolution images and has many characteristics similar with object detection. For instance, semantic segmentation also requires a larger input resolution and window size, and multi-resolution feature maps for visual recognition at a finer granularity. On these two points, we obtain similar observations for semantic segmentation as with object detection: larger input with unchanged window size achieves relatively good performance, and directly using the output features from different stages of Swin Transformer as the input of the head network also works well for semantic segmentation. For this task, we adopt the most widely-used benchmark, ADE20K~\cite{zhou2018ade}, for evaluation. ADE20K covers a broad range of 150 semantic categories. It has 25K images in total, with 20K for training, 2K for validation, and another 3K for testing. 

\paragraph{Video Action Recognition} The main challenge of adapting the frozen pretrained image model to video action recognition lies in the different input formats. Video action recognition aims to recognize the action types of each input video, which consists of a sequence of video frames (e.g., 16 frames per clip). Therefore, it is essential in video action recognition to capture both spatial and temporal relationships. Previous works based on full finetuning mainly focus on exploring the simultaneous modeling of spatial and temporal relationships in the base network. For the frozen setting, however, the pretraining model is capable only of spatial reasoning, so the head network must compensate for the lack of temporal modeling. For the task of human action recognition, we adopt the widely-used Kinetics-400 \cite{kay2017kinetics} dataset, consisting of $\sim$240k training videos and 20k validation videos over 400 human action categories. 

\section{Which Pretraining Task Fits Best with the Frozen Setting?}
\label{sec:which_task_fit}

In this section, we evaluate four prevalent pretraining tasks under the frozen setting, namely supervised pretraining, contrastive learning, masked image modeling, and visual-linguistic learning. Specifically, we use six pretrained Swin Transformer models, which include supervised pretraining on ImageNet-1K (SUP-1K), supervised pretraining on ImageNet-22K (SUP-22K), contrastive learning of EsViT on ImageNet-1K (EsViT-1K), masked image modeling of SimMIM on ImageNet-1K (SimMIM-1K), image-text alignment of iCar on Laion\cite{schuhmann2021laion}(iCar-Laion), and jointly training of image-text alignment and image classification for iCar on Laion and ImageNet-22K (iCar-Laion-22K). For the downstream tasks, we adopt SwinV2-B~\cite{swinv2} as the base network and three widely-used frameworks as the head networks: Mask R-CNN~\cite{Mask-rcnn} with FPN~\cite{FPN} for COCO object detection, Mask2Former~\cite{mask2former} with a one-block pixel decoder for ADE20K semantic segmentation, and a 
spatial-only Video-Swin-Transformer~\cite{liu2021video} variant, where a temporal window size of 1 is used, with a linear head for Kinetics-400 action recognition.

\begin{table}[htb]
        \centering
        \addtolength{\tabcolsep}{-1.0pt}
        \begin{tabular}{c|cccc|cc|cc}
            \Xhline{1.0pt}
            \multirow{3}{*}{Approach} & \multicolumn{4}{c|}{COCO} & \multicolumn{2}{c|}{ADE20K} & \multicolumn{2}{c}{Kinetics-400}\\
            & \multicolumn{2}{c}{Frozen}& \multicolumn{2}{c|}{Full ft.} & Frozen & Full ft. & Frozen & Full ft. \\
            & AP$^\text{box}$ & AP$^\text{mask}$ & AP$^\text{box}$ & AP$^\text{mask}$ & mIoU & mIoU & acc@1 & acc@1 \\
            \hline
            SUP-1K & 42.4 & 38.7 & 50.5 & 44.5 & 49.8 & 52.3 & 60.4 & 77.0 \\
            SUP-22K & 45.0 & 41.1 & 51.9 & 45.7 & 51.9 & 55.3 & 70.3 & 79.7 \\
            EsViT-1K & 42.0 & 38.5 & 51.5 & 45.6 & 49.7 & 52.1 & 62.0 & 76.5 \\
            SimMIM-1K & 34.1 & 32.4 & 52.9 & 46.7 & 42.4 & 51.7 & 14.2 & 75.9 \\
            iCAR-Laion & 43.3 & 39.5 & 51.7 & 45.5 & 51.0 & 55.3 & 65.1 & 79.5 \\
            iCAR-Laion-22K & 44.9 & 41.2 & 52.3 & 46.1 & 51.1 & 55.4 & 69.4 & 80.2 \\
          \Xhline{1.0pt}  
       \end{tabular}
     \caption{Comparison of different pretraining tasks on frozen and full finetuning settings. Results of box mAP and mask mAP for COCO object detection, mIoU for ADE20K semantic segmentation, and top-1 accuracy for Kinetics-400 action recognition are reported. SUP denotes supervised classification as pretraining.}
     \label{tab:pretrain_task}
\end{table}

Results are shown in Table~\ref{tab:pretrain_task}. For the models pretrained on the ImageNet-1K dataset, we can observe that SUP-1K and EsViT-1K perform similarly on almost all the benchmarks, including both the full finetuning and frozen settings. This phenomenon is understandable, because the linear evaluation result of EsViT is relatively high, indicating that the features extracted by EsViT-1K are similar to the SUP-1K counterpart. The SimMIM pretrained model performs competitively high in the full finetuning setting. For example, it achieves much better performance on COCO, comparable  performance on ADE, and slightly worse performance on K400, in comparison to the SUP-1K counterpart. However, the SimMIM-1K model performs poorly on all benchmarks in the frozen setting. This could be foreseen from its poor linear evaluation performance on ImageNet-1K, indicating that its output features do not capture high-level semantics. The iCAR-Laion model outperforms SUP-1K and EsViT-1K on almost all the benchmarks, indicating the effectiveness of learning from large scale image-text datasets. In addition, with jointly training on Laion and ImageNet-22K, the iCAR-Laion-22K model achieves similar results with the SUP-22K model. We can also observe that the SUP-22K model outperforms SUP-1K significantly in all settings, reflecting the great benefit that data scaling brings to supervised pretraining. In general, we find that supervised pretraining works best under the frozen setting, and we adopt supervised pretrained models by default in the following experiments.  However, in all three downstream tasks, the frozen settings still underperform full finetuning by substantial margins. 

\section{What is the Key to Making the Frozen Setting Work?}
\label{sec:key_to_work}

Towards improving performance over disparate tasks, in this section we consider how to extend the frozen setting for better task adaptation. 
For the following investigation, we utilize the best-performing pretraining task, supervised classification on ImageNet-22K (SUP-22K).

\subsection{Adding More Tunable Parameters at Head Networks} 
To some extent, the poor performance of the frozen setting is understandable. Despite the significant difference between the pretraining and downstream task, all parameters in the base networks are locked in the frozen setting, and only a few newly-added parameters are available for task adaptation. Thus, a direct solution is to add more tunable parameters to the task-specific head networks. As the head network varies among different downstream tasks, we present task-specific strategies on how to add parameters to the head.

\begin{table}[h]
        \centering
        \addtolength{\tabcolsep}{3.0pt}
        \begin{tabular}{c|cc|cc}
            \Xhline{1.0pt}
            \multirow{2}{*}{Head Network} & \multicolumn{2}{c|}{Frozen} & \multicolumn{2}{c}{Full ft.} \\
            & AP$^\text{box}$ & AP$^\text{mask}$ & AP$^\text{box}$ & AP$^\text{mask}$ \\
            \hline
            FPN & 45.0 & 41.1 & 51.9 & 45.7  \\
            FPN w. $4\times$ Residual Blocks inside & 49.5 & 44.0 & 52.1 & 45.6 \\
            BiFPN & 51.9 & 46.0 & 52.3 & 45.7 \\
            FPN w. Cascade Head & 49.0 & 43.0 & 54.5 & 46.9 \\
            BiFPN w. Cascade Head & 53.8 & 46.7 & 54.3 & 46.9 \\
          \Xhline{1.0pt}  
       \end{tabular}
       \caption{Results for different head networks on COCO object detection and instance segmentation, including FPN with vanilla box/mask head, FPN with four more residual blocks inside, BiFPN, FPN with cascade head, and BiFPN with cascade head.}
    \label{tab:param_at_head_coco}
\end{table}

\paragraph{Object Detection} For object detection, we adopt the framework of Mask R-CNN~\cite{Mask-rcnn} with FPN~\cite{FPN} as the head network. Previous work on Mask R-CNN mainly explored improvements over FPN and the box/mask head, so we also examine common alternatives to these two components, e.g., replacing FPN with BiFPN~\cite{tan2020efficientdet} or adding more residual blocks in FPN, and changing the vanilla box/mask head to a cascaded box/mask head~\cite{cai2018cascade}. Results are shown in Table~\ref{tab:param_at_head_coco}. 
We can observe that adding parameters to FPN, e.g., replacing FPN with BiFPN or adding more residual blocks, both significantly bridge the gap between the frozen setting and full finetuning, from -6.9 to -2.6 box mAP and -6.9 to -0.4 box mAP, respectively. As BiFPN is carefully designed with many interactions between multi-resolution feature maps, replacing FPN with BiFPN outperforms the model of adding more residual blocks in FPN. However, replacing the box/mask head with a cascade head~\cite{cai2018cascade} improves the performance of both the frozen and full finetuning settings, from 45.0 to 49.0 box mAP and 51.9 to 54.5 box mAP, but does not bridge the gap between them. 
Adding more tunable parameters to FPN is thus found to be more effective than adding to the box/mask head.

\begin{figure}[h]
    \centering
    \includegraphics[width=\linewidth]{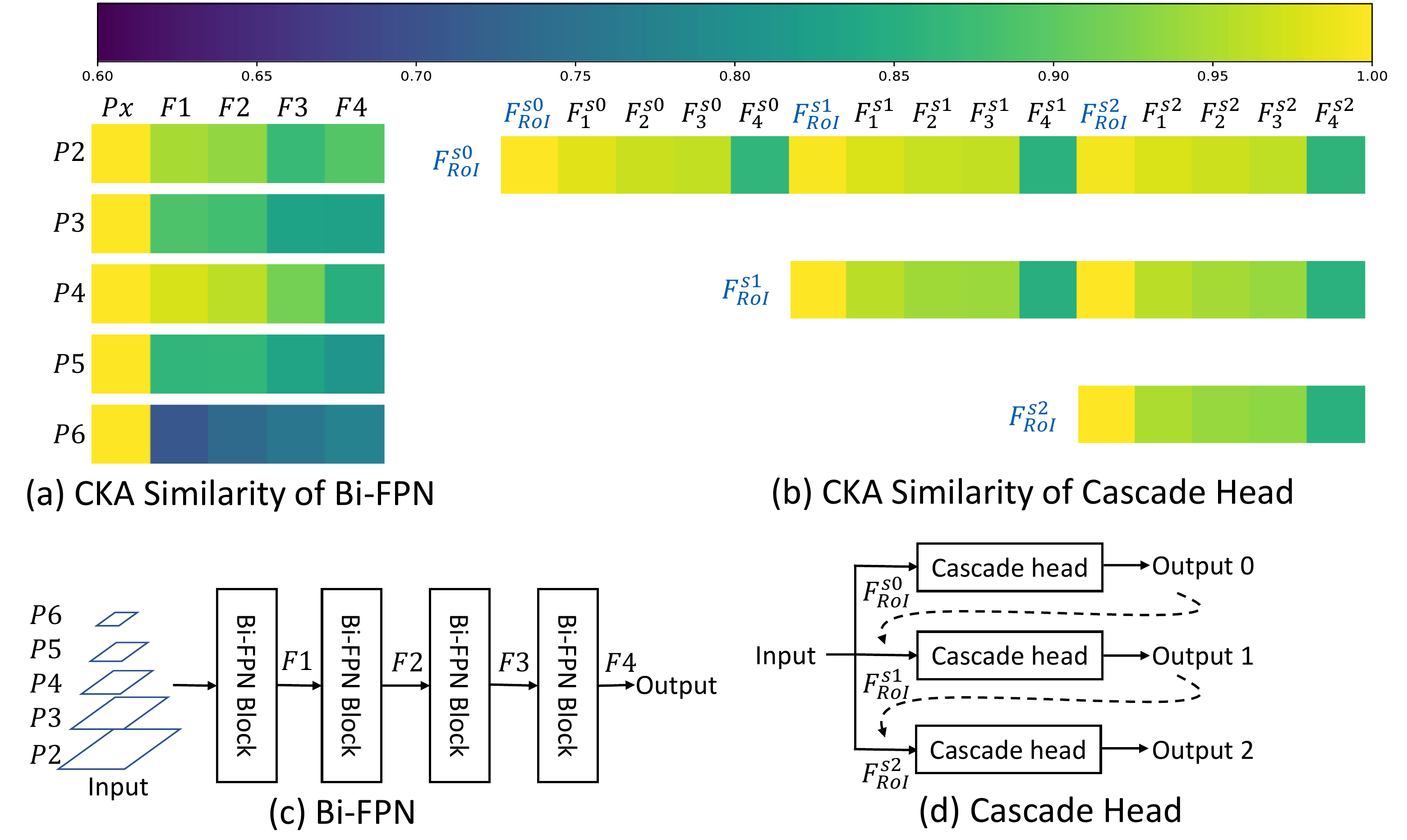}
    \caption{The CKA heatmap on the features across layers in BiFPN (a) and Cascade Head (b). The detailed architecture of BiFPN (c) and Cascade Head (d).}
    \label{fig:cka}
\end{figure}

To understand this difference in behavior, we conduct a performance analysis and an examination of feature similarity across different layers in BiFPN and Cascade Head by Centered Kernel Alignment (CKA)~\cite{cka}.
As shown in Figure~\ref{fig:cka} (a), for each stage (with different resolutions of feature map) of BiFPN, we plot the CKA similarity between the input features ($Px$ of stage $x$) and the output of each block ($Fx$ of block $x$).
For the cascade head, as shown in Figure~\ref{fig:cka} (b), we plot the CKA similarity between the input features of each stage ($F_{RoI}^{si}$ in stage $i$) and the hidden features inside each stage ($F_{j}^{si}$: output of $j$'s block in stage $i$).
From this figure, we can observe that the features across layers and even stages in cascade heads are almost the same except for the last output, but in BiFPN, features across different layers are different (with lower CKA similarity). In other words, adding parameters in BiFPN provide more useful capacity for transformation, but adding parameters in the cascade head hardly provides useful computation. 
This difference can be explained by the architecture of BiFPN (Figure~\ref{fig:cka}(c)) and cascade head (Figure~\ref{fig:cka}(d)). BiFPN follows a sequential structure, where the input of each block is the output of the previous block. The cascade head follows a parallel structure, and all stages extract RoI features from the original multi-resolution feature map, making the input of each stage have extremely high CKA similarity. Also, the transformations made by each stage are similar.
The last output of the cascade head behaves differently because it is followed by a pooling layer, which is proven to have a great impact on CKA similarity. Consequently, the number of well-placed tunable parameters is the key to make frozen settings work well.

\begin{table}[h]
               \centering
        \addtolength{\tabcolsep}{3.0pt}
        \begin{tabular}{c|cc}
            \Xhline{1.0pt}
            Head Network& \makecell{Frozen\\mIoU} & \makecell{Full Ft.\\mIoU}\\
            \hline
            $1\times$ Pixel Dec. w. ($3$+$1$)$\times$ Transformer Dec. & 51.0 & 55.4 \\
            $6\times$ Pixel Dec. w. ($3$+$1$)$\times$ Transformer Dec.  & 52.6 & 55.3 \\
            $1\times$ Pixel Dec. w. ($9$+$1$)$\times$ Transformer Dec. & 51.9 & 55.3 \\
            $6\times$ Pixel Dec. w. ($9$+$1$)$\times$ Transformer Dec.  & 53.2 & 55.5 \\
          \Xhline{1.0pt}  
       \end{tabular}
       \caption{Results from using different head networks on ADE20K semantic segmentation, with either one or six pixel decoder blocks, and a Transformer decoder with four or ten blocks.}
    \label{tab:param_at_head_ade}
\end{table}

\paragraph{Semantic Segmentation}
For semantic segmentation, we adopt Mask2former~\cite{mask2former} with a one-block pixel decoder and four-block transformer decoder as the head network. The framework of Mask2former is similar to Mask R-CNN in object detection, where the pixel decoder corresponds to the FPN, and the transformer head corresponds to the box/mask head. Thus, we employ a setting similar to that used for object detection, by increasing the number of parameters in these two major components, e.g., changing the pixel decoder to six blocks and the transformer decoder to ten blocks. As shown in Table~\ref{tab:param_at_head_ade}, similar results can be observed where enlarging the pixel decoder helps more in bridging the gap than enlarging the transformer head. We also perform a CKA analysis on this framework and observe behaviors like that in object detection, as shown in the Appendix.

\begin{table}[h]
        \centering
        \addtolength{\tabcolsep}{3.0pt}
        \begin{tabular}{c|cc}
            \Xhline{1.0pt}
            Head Network& \makecell{Frozen\\top-1 accuracy} & \makecell{Full Ft.\\top-1 accuracy}\\
            \hline
            Linear & 70.3 & 79.7 \\
            $4\times$ Temporal Blocks & 74.7 & 79.5 \\
            $4\times$ Global Blocks & 77.5 & 79.8 \\
          \Xhline{1.0pt}  
       \end{tabular}
       \caption{Results with different head networks on Kinetics-400 video action recognition: linear classifier, four temporal Transformer blocks, and four global (spatiotemporal) Transformer blocks.}
    \label{tab:param_at_head_k400}
\end{table}

\paragraph{Video Action Recognition}
For video action recognition, it is crucial to capture both spatial and temporal relationships, but the pretraining model is only capable of spatial reasoning. Under the frozen setting, the lack of temporal modeling thus needs to be compensated for in the head network. In our experiments, we adopt a spatial-then-temporal Video-Swin-Transformer~\cite{liu2021video} framework that is enhanced with additional Transformer blocks, either temporal-only or global (spatiotemporal). In Table~\ref{tab:param_at_head_k400}, we can observe that under the full finetuning setting, adding global blocks and adding temporal blocks work similarly with a linear classifier. On the other hand, adding temporal blocks in the frozen setting leads to significantly better performance than the linear classifier setting, which verifies the need for temporal reasoning. Also, adding global blocks in the frozen setting improves substantially over adding temporal-only blocks. Although the base network is pretrained for spatial reasoning and global blocks are added, there is still a non-trivial gap between the frozen and full finetuning settings. This may be due to some missing information from the original features, and the remaining gap may be reduced by further introducing a network with independent access to the input, which is beyond the scope of this paper and left for future work.

\begin{figure}[h]
    \centering
    \begin{subfigure}[b]{0.32\textwidth}
        \centering
        \includegraphics[width=\textwidth]{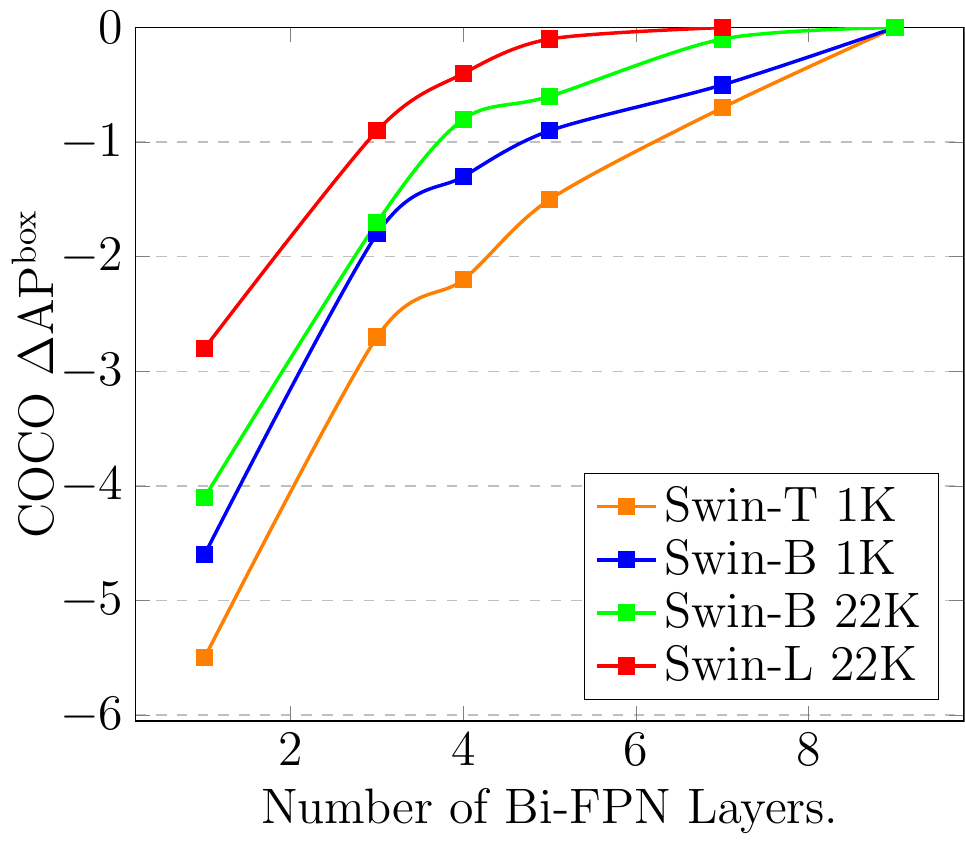}
        \caption{COCO}
    \end{subfigure}
    \hfill
    \begin{subfigure}[b]{0.32\textwidth}
        \centering
        \includegraphics[width=\textwidth]{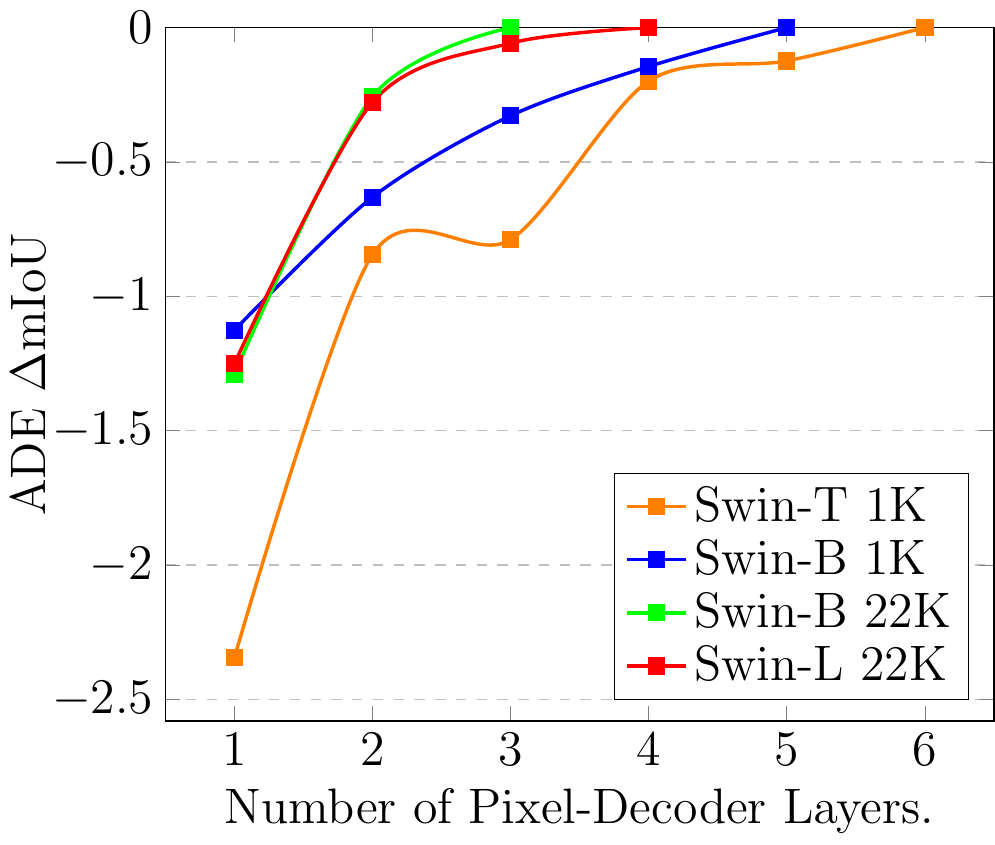}
        \caption{ADE20K}
    \end{subfigure}
    \hfill
    \begin{subfigure}[b]{0.32\textwidth}
        \centering
        \includegraphics[width=\textwidth]{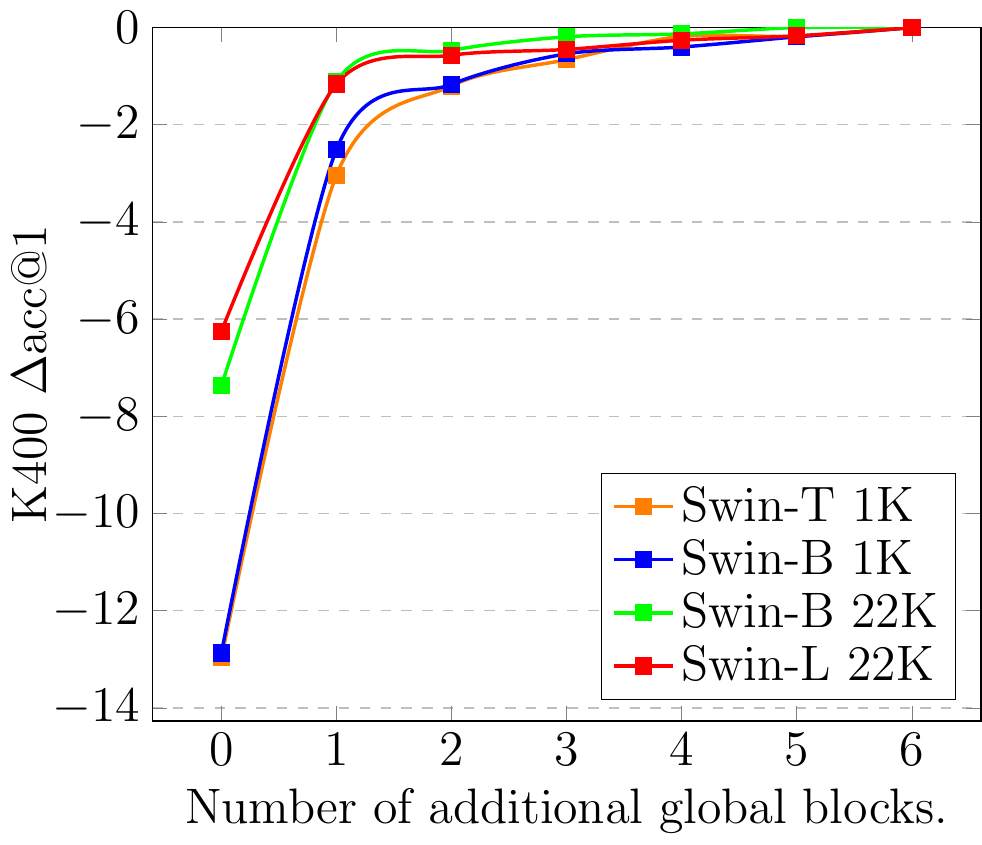}
        \caption{Kinetics-400}
    \end{subfigure}
    \caption{Performance improvements on COCO object detection, ADE semantic segmentation, and Kinetics-400 action recognition with increases in the number of trainable parameters.}
    \label{fig:param_head}
\end{figure}

\subsection{Does Larger Pretrained Base Network Need Smaller Head Network?} 
When the pretrained model becomes larger, it generally contains richer and more valuable information, so for downstream tasks, we generally expect that a larger pretrained model does not require as large of a head. In NLP, it is verified that only a few layers or a few prompts are needed when the pretrained model is extremely large, e.g., 173B. For computer vision, we seek an answer to whether the current best pretrained models have similar properties. We perform the empirical study with different sizes of pretrained models, SwinV2-T/B with ImageNet-1K pretraining, and SwinV2-B/L with ImageNet-22K pretraining, on the three representative downstream tasks. 
From Figure~\ref{fig:param_head} (a), on COCO object detection, we can surprisingly observe a clear trend, that SwinV2-L-22K requires fewer parameters to converge (a 5-layer BiFPN), while SwinV2-T-1K requires more parameters to converge (a 9-layer BiFPN).
On ADE20K semantic segmentation in Figure~\ref{fig:param_head} (b), 
we also find that SwinV2-T-1K requires more parameters to converge (a 6-layer pixel decoder) than other larger base networks.
On Kinetics-400 video action recognition in Figure~\ref{fig:param_head} (c), different models appear to require a similar number of parameters to converge. This observation matches the previous one, that after careful tuning, there is still a non-trivial gap between the frozen and full finetuning settings on video action recognition (as shown in Table~\ref{tab:param_at_head_k400}), indicating a larger difference between pretraining and video recognition tasks than others.

\section{Results with Different Sizes of Base Networks}
\label{sec:different_size_result}

\begin{table}[htb]
        \centering
        \addtolength{\tabcolsep}{-2.5pt}
        \begin{tabular}{c|ccccc|cc|cc}
            \Xhline{1.0pt}
            \multirow{3}{*}{Base Network} & \multicolumn{5}{c|}{COCO} & \multicolumn{2}{c|}{ADE20K} & \multicolumn{2}{c}{Kinetics-400}\\
            & {Head} &  \multicolumn{2}{c}{Frozen}& \multicolumn{2}{c|}{Full ft.} & Frozen & Full ft. & Frozen & Full ft. \\
            & Network &  AP$^\text{box}$ & AP$^\text{mask}$ & AP$^\text{box}$ & AP$^\text{mask}$ & mIoU & mIoU & acc@1 & acc@1 \\
            \hline
            SwinV2-T-1K & Mask.  & 49.6 & 44.0 & 49.8 & 43.8 & 48.9 & 50.3 &  70.6 & 75.7  \\
            SwinV2-B-1K & Mask. & 50.3 & 44.5 & 51.1 & 44.8 & 51.5 & 52.7 & 73.3 & 77.9  \\
            SwinV2-B-22K & Mask. & 52.5 & 46.6 & 52.3 & 45.7 & 53.2 & 55.5 & 77.7 & 79.8 \\
            SwinV2-L-22K & Mask.  & 53.0 & 46.8 & 53.0 & 46.5 & 54.4 & 56.5 & 78.7 & 79.8 \\
            \hline
            SwinV2-B-22K & Cascade.  & 53.8 & 46.7 & 54.3 & 46.9 &- & - & -&- \\
            \multirow{2}{*}{SwinV2-G-ext22K} & HTC  & 57.9 & 50.4 & - & - & \multirow{2}{*}{57.6} & \multirow{2}{*}{-}& \multirow{2}{*}{81.7} & \multirow{2}{*}{-} \\
             & HTC$^\dag$  & 59.3 & 51.6 & - & - &  & &  & \\
          \Xhline{1.0pt}  
       \end{tabular}
     \caption{Comparison of different-sized base networks with carefully designed head networks on frozen and full finetuning settings, including SwinV2-T/B pretrained on ImageNet-1K,  SwinV2-B/L pretrained on ImageNet-22K, and SwinV2-G pretrained on ext22K. Results of box mAP and mask mAP for COCO object detection validation set, mIoU for ADE20K semantic segmentation, and top-1 accuracy for Kinetics-400 action recognition are reported. Mask. denotes Mask R-CNN is used as the head network, and Cascade. refers to Cascade Mask R-CNN. $^\dag$ denotes multi-scale testing on COCO.}
     \label{tab:3b}
\end{table}

After careful design of the head networks for each downstream task, we compare different sizes of base network with the best setup of frozen and full finetuning settings, including SwinV2-T/B pretrained on ImageNet-1K, and SwinV2-B/L pretrained on ImageNet-22K, shown in Table~\ref{tab:3b}. On COCO object detection, the performance gap is well bridged, where SwinV2-B/L-22K even achieve on-par performance under the frozen and full finetuning settings with two different head networks. But on ADE20K, there is still a significant performance gap. For example, using SwinV2-L-22K as the base network, the frozen setting still underperforms the full finetuning setting by $2.1$ mIoU. A large performance gap between the frozen and full finetuning settings can also be observed on Kinetics-400 video action recognition. Interestingly, the performance gaps on ImageNet-1K pretrained models are significantly larger than that for models pretrained on ImageNet-22K, indicating that enlarging the pretraining dataset from ImageNet-1K to ImageNet-22K helps to bridge the task gap of pretraining and finetuning on action recognition.

We additionally explore the upper bound of performance under the frozen setting, using a frozen supervised pretrained model with 3 billion parameters trained on 70M images (SwinV2-G), shown in Table~\ref{tab:3b}. With this giant model, highly competitive performance is achieved on major benchmarks: 59.3/60.0 box mAP and 51.6/52.2 mask mAP on COCO validation/test-dev sets, 57.6 mIoU on ADE20K semantic segmentation, and 81.7 top-1 accuracy on Kinetics-400 action recognition. 

\section{When Does Frozen Setting Outperform Full Finetuning?}

To further explore the potential of the frozen setting, we compare it with full finetuning in the low-data scenario. More specifically, we test these two settings with a small portion of the COCO training dataset. Following the standard practice in semi-supervised learning~\cite{sohn2020simple_semi, xu2021softteacher}, we construct two datasets with only 1\% and 10\% (about 12K and 120K training images, respectively) of COCO. 

\begin{table}[h]
        \centering
        \begin{tabular}{c|cccc|cccc}
            \Xhline{1.0pt}
            \multirow{3}{*}{Head Network} &  \multicolumn{4}{c|}{1\% data} & \multicolumn{4}{c}{10\% data} \\
            & \multicolumn{2}{c}{Frozen} & \multicolumn{2}{c|}{Full ft.} & \multicolumn{2}{c}{Frozen} & \multicolumn{2}{c}{Full ft.} \\
            & AP$^\text{box}$ & AP$^\text{mask}$ & AP$^\text{box}$ & AP$^\text{mask}$ & AP$^\text{box}$ & AP$^\text{mask}$ & AP$^\text{box}$ & AP$^\text{mask}$ \\
            \hline
            FPN & 27.5 & 26.4 & 27.5 & 25.8 & 39.1 & 36.2 & 42.2 & 37.5 \\
            $1\times$ BiFPN Layer & 29.8 & 27.6 & 26.3 & 24.7 & 42.8 & 38.5 & 42.0 & 37.3 \\
            $2\times$ BiFPN Layer & 31.4 & 28.4 & 26.3 & 24.6 & 44.2 & 39.4  & 42.1 & 37.3 \\
            $3\times$ BiFPN Layer & 32.1 & 28.9 & 25.9 & 24.3 & 44.6 & 39.7 & 42.3 & 37.6 \\
            $4\times$ BiFPN Layer & 31.9 & 28.6 & 26.0 & 24.1 & 45.0 & 40.0 & 42.1 & 37.4 \\
          \Xhline{1.0pt}  
       \end{tabular}
       \caption{Results for different head networks on COCO with 1\% and 10\% labeled data. Head networks include FPN and BiFPN.}
    \label{tab:semi_coco}
\end{table}

Results are shown in Table \ref{tab:semi_coco}. For training with 1\% labeled data, when adopting FPN as head, the frozen setting achieves performance competitive with the full finetuning one. With an increasing number of BiFPN layers, we can observe stable performance improvements for the frozen setting and inferior performance for full finetuning, which indicates an overfitting issue for the full finetuning setting.
For training with 10\% data, there is a performance gap of -3.1 box mAP between the frozen setting and full finetuning with FPN as the head network. Compared with training on the full COCO training dataset, this gap is reduced from -6.9 box mAP to -3.1 box mAP. When replacing FPN with BiFPN, the frozen setting surpasses full finetuning with a clear margin of 2.9 box AP and 2.6 mask AP.

\section{Conclusion}
In this paper, we presented a detailed empirical study of frozen pretrained models for diverse computer vision tasks. Based on our observations, we proposed strategies for extending the frozen setting to work effectively for various downstream tasks, and also found that a universal representation with solid performance over assorted benchmarks can be learned for a giant frozen model. Our study highlights the strong potential of this transfer learning approach, and we hope it will kindle further interest in this research direction.

\newpage

\bibliographystyle{apalike}
\bibliography{ref}

\begin{thebibliography}{}

\bibitem[Bao et~al., 2021]{bao2021beit}
Bao, H., Dong, L., and Wei, F. (2021).
\newblock Beit: Bert pre-training of image transformers.
\newblock {\em arXiv preprint arXiv:2106.08254}.

\bibitem[Bilen and Vedaldi, 2017]{bilen2017universal}
Bilen, H. and Vedaldi, A. (2017).
\newblock Universal representations: The missing link between faces, text,
  planktons, and cat breeds.
\newblock {\em arXiv preprint arXiv:1701.07275}.

\bibitem[Brown et~al., 2020]{gpt3}
Brown, T., Mann, B., Ryder, N., Subbiah, M., Kaplan, J.~D., Dhariwal, P.,
  Neelakantan, A., Shyam, P., Sastry, G., Askell, A., et~al. (2020).
\newblock Language models are few-shot learners.
\newblock {\em Advances in neural information processing systems},
  33:1877--1901.

\bibitem[Cai and Vasconcelos, 2018]{cai2018cascade}
Cai, Z. and Vasconcelos, N. (2018).
\newblock Cascade r-cnn: Delving into high quality object detection.
\newblock In {\em Proceedings of the IEEE conference on computer vision and
  pattern recognition}, pages 6154--6162.

\bibitem[Cao et~al., 2020]{cao2020pic}
Cao, Y., Xie, Z., Liu, B., Lin, Y., Zhang, Z., and Hu, H. (2020).
\newblock Parametric instance classification for unsupervised visual feature
  learning.
\newblock {\em Advances in Neural Information Processing Systems}, 33.

\bibitem[Carreira and Zisserman, 2017]{carreira2017quo}
Carreira, J. and Zisserman, A. (2017).
\newblock Quo vadis, action recognition? a new model and the kinetics dataset.
\newblock In {\em proceedings of the IEEE Conference on Computer Vision and
  Pattern Recognition}, pages 6299--6308.

\bibitem[Chen et~al., 2019]{chen2019htc}
Chen, K., Pang, J., Wang, J., Xiong, Y., Li, X., Sun, S., Feng, W., Liu, Z.,
  Shi, J., Ouyang, W., et~al. (2019).
\newblock Hybrid task cascade for instance segmentation.
\newblock In {\em Proceedings of the IEEE/CVF Conference on Computer Vision and
  Pattern Recognition}, pages 4974--4983.

\bibitem[Chen et~al., 2020a]{chen2020imagegpt}
Chen, M., Radford, A., Child, R., Wu, J., and Jun, H. (2020a).
\newblock Generative pretraining from pixels.
\newblock {\em Advances in Neural Information Processing Systems}.

\bibitem[Chen et~al., 2020b]{chen2020simclr}
Chen, T., Kornblith, S., Norouzi, M., and Hinton, G. (2020b).
\newblock A simple framework for contrastive learning of visual
  representations.
\newblock {\em ICML}.

\bibitem[Cheng et~al., 2021]{mask2former}
Cheng, B., Misra, I., Schwing, A.~G., Kirillov, A., and Girdhar, R. (2021).
\newblock Masked-attention mask transformer for universal image segmentation.
\newblock {\em arXiv preprint arXiv:2112.01527}.

\bibitem[Deng et~al., 2009]{deng2009imagenet}
Deng, J., Dong, W., Socher, R., Li, L.-J., Li, K., and Fei-Fei, L. (2009).
\newblock Imagenet: A large-scale hierarchical image database.
\newblock In {\em CVPR}, pages 248--255. Ieee.

\bibitem[Devlin et~al., 2018]{devlin2018bert}
Devlin, J., Chang, M.-W., Lee, K., and Toutanova, K. (2018).
\newblock Bert: Pre-training of deep bidirectional transformers for language
  understanding.
\newblock {\em arXiv preprint arXiv:1810.04805}.

\bibitem[Donahue et~al., 2014]{donahue2014decaf}
Donahue, J., Jia, Y., Vinyals, O., Hoffman, J., Zhang, N., Tzeng, E., and
  Darrell, T. (2014).
\newblock Decaf: A deep convolutional activation feature for generic visual
  recognition.
\newblock In {\em International conference on machine learning}, pages
  647--655. PMLR.

\bibitem[Dosovitskiy et~al., 2021]{dosovitskiy2020vit}
Dosovitskiy, A., Beyer, L., Kolesnikov, A., Weissenborn, D., Zhai, X.,
  Unterthiner, T., Dehghani, M., Minderer, M., Heigold, G., Gelly, S.,
  Uszkoreit, J., and Houlsby, N. (2021).
\newblock An image is worth 16x16 words: Transformers for image recognition at
  scale.
\newblock In {\em International Conference on Learning Representations}.

\bibitem[Dosovitskiy et~al., 2014]{dosovitskiy2014exemplarcnn}
Dosovitskiy, A., Springenberg, J.~T., Riedmiller, M., and Brox, T. (2014).
\newblock Discriminative unsupervised feature learning with convolutional
  neural networks.
\newblock In {\em Advances in neural information processing systems}, pages
  766--774.

\bibitem[Ghiasi et~al., 2021]{simple_copy_paste}
Ghiasi, G., Cui, Y., Srinivas, A., Qian, R., Lin, T.-Y., Cubuk, E.~D., Le,
  Q.~V., and Zoph, B. (2021).
\newblock Simple copy-paste is a strong data augmentation method for instance
  segmentation.
\newblock In {\em Proceedings of the IEEE/CVF Conference on Computer Vision and
  Pattern Recognition (CVPR)}, pages 2918--2928.

\bibitem[Girshick et~al., 2014]{girshick2014rich}
Girshick, R., Donahue, J., Darrell, T., and Malik, J. (2014).
\newblock Rich feature hierarchies for accurate object detection and semantic
  segmentation.
\newblock In {\em Proceedings of the IEEE conference on computer vision and
  pattern recognition}, pages 580--587.

\bibitem[Grill et~al., 2020]{grill2020byol}
Grill, J.-B., Strub, F., Altch{\'e}, F., Tallec, C., Richemond, P.,
  Buchatskaya, E., Doersch, C., Avila~Pires, B., Guo, Z., Gheshlaghi~Azar, M.,
  et~al. (2020).
\newblock Bootstrap your own latent-a new approach to self-supervised learning.
\newblock {\em Advances in Neural Information Processing Systems}, 33.

\bibitem[He et~al., 2021]{he2021masked}
He, K., Chen, X., Xie, S., Li, Y., Doll{\'a}r, P., and Girshick, R. (2021).
\newblock Masked autoencoders are scalable vision learners.
\newblock {\em arXiv preprint arXiv:2111.06377}.

\bibitem[He et~al., 2020]{he2019moco}
He, K., Fan, H., Wu, Y., Xie, S., and Girshick, R. (2020).
\newblock Momentum contrast for unsupervised visual representation learning.
\newblock {\em CVPR}.

\bibitem[He et~al., 2017]{Mask-rcnn}
He, K., Gkioxari, G., Doll{\'a}r, P., and Girshick, R. (2017).
\newblock Mask r-cnn.
\newblock In {\em ICCV}, pages 2961--2969.

\bibitem[He et~al., 2016]{he2016resnet}
He, K., Zhang, X., Ren, S., and Sun, J. (2016).
\newblock Deep residual learning for image recognition.
\newblock In {\em CVPR}, pages 770--778.

\bibitem[Houlsby et~al., 2019]{houlsby2019parameter}
Houlsby, N., Giurgiu, A., Jastrzebski, S., Morrone, B., De~Laroussilhe, Q.,
  Gesmundo, A., Attariyan, M., and Gelly, S. (2019).
\newblock Parameter-efficient transfer learning for nlp.
\newblock In {\em International Conference on Machine Learning}, pages
  2790--2799. PMLR.

\bibitem[Hu et~al., 2021]{hu2021lora}
Hu, E.~J., Shen, Y., Wallis, P., Allen-Zhu, Z., Li, Y., Wang, S., Wang, L., and
  Chen, W. (2021).
\newblock Lora: Low-rank adaptation of large language models.
\newblock {\em arXiv preprint arXiv:2106.09685}.

\bibitem[Jia et~al., 2021]{jia2021align}
Jia, C., Yang, Y., Xia, Y., Chen, Y.-T., Parekh, Z., Pham, H., Le, Q., Sung,
  Y.-H., Li, Z., and Duerig, T. (2021).
\newblock Scaling up visual and vision-language representation learning with
  noisy text supervision.
\newblock In {\em International Conference on Machine Learning}, pages
  4904--4916. PMLR.

\bibitem[Kay et~al., 2017]{kay2017kinetics}
Kay, W., Carreira, J., Simonyan, K., Zhang, B., Hillier, C., Vijayanarasimhan,
  S., Viola, F., Green, T., Back, T., Natsev, P., et~al. (2017).
\newblock The kinetics human action video dataset.
\newblock {\em arXiv preprint arXiv:1705.06950}.

\bibitem[Kolesnikov et~al., 2019]{alex2019big}
Kolesnikov, A., Beyer, L., Zhai, X., Puigcerver, J., Yung, J., Gelly, S., and
  Houlsby, N. (2019).
\newblock Big transfer (bit): General visual representation learning.

\bibitem[Kornblith et~al., 2019a]{cka}
Kornblith, S., Norouzi, M., Lee, H., and Hinton, G. (2019a).
\newblock Similarity of neural network representations revisited.
\newblock In {\em International Conference on Machine Learning}, pages
  3519--3529. PMLR.

\bibitem[Kornblith et~al., 2019b]{kornblith2019better}
Kornblith, S., Shlens, J., and Le, Q.~V. (2019b).
\newblock Do better imagenet models transfer better?
\newblock In {\em Proceedings of the IEEE/CVF conference on computer vision and
  pattern recognition}, pages 2661--2671.

\bibitem[Krizhevsky et~al., 2012]{alexnet}
Krizhevsky, A., Sutskever, I., and Hinton, G.~E. (2012).
\newblock Imagenet classification with deep convolutional neural networks.
\newblock In {\em Advances in Neural Information Processing Systems}, pages
  1097--1105.

\bibitem[Lester et~al., 2021]{lester2021power}
Lester, B., Al-Rfou, R., and Constant, N. (2021).
\newblock The power of scale for parameter-efficient prompt tuning.
\newblock {\em arXiv preprint arXiv:2104.08691}.

\bibitem[Li et~al., 2021]{esvit}
Li, C., Yang, J., Zhang, P., Gao, M., Xiao, B., Dai, X., Yuan, L., and Gao, J.
  (2021).
\newblock Efficient self-supervised vision transformers for representation
  learning.
\newblock {\em arXiv preprint arXiv:2106.09785}.

\bibitem[Li and Liang, 2021]{li2021prefix}
Li, X.~L. and Liang, P. (2021).
\newblock Prefix-tuning: Optimizing continuous prompts for generation.
\newblock {\em arXiv preprint arXiv:2101.00190}.

\bibitem[Lin et~al., 2017]{FPN}
Lin, T.-Y., Doll{\'a}r, P., Girshick, R., He, K., Hariharan, B., and Belongie,
  S. (2017).
\newblock Feature pyramid networks for object detection.
\newblock In {\em ICCV}, pages 2117--2125.

\bibitem[Lin et~al., 2014]{lin2014coco}
Lin, T.-Y., Maire, M., Belongie, S., Hays, J., Perona, P., Ramanan, D.,
  Doll{\'a}r, P., and Zitnick, C.~L. (2014).
\newblock Microsoft coco: Common objects in context.
\newblock In {\em European conference on computer vision}, pages 740--755.
  Springer.

\bibitem[Liu et~al., 2021a]{liu2021pre}
Liu, P., Yuan, W., Fu, J., Jiang, Z., Hayashi, H., and Neubig, G. (2021a).
\newblock Pre-train, prompt, and predict: A systematic survey of prompting
  methods in natural language processing.
\newblock {\em arXiv preprint arXiv:2107.13586}.

\bibitem[Liu et~al., 2021b]{liu2021ptuning}
Liu, X., Zheng, Y., Du, Z., Ding, M., Qian, Y., Yang, Z., and Tang, J. (2021b).
\newblock Gpt understands, too.
\newblock {\em arXiv preprint arXiv:2103.10385}.

\bibitem[Liu et~al., 2021c]{swinv2}
Liu, Z., Hu, H., Lin, Y., Yao, Z., Xie, Z., Wei, Y., Ning, J., Cao, Y., Zhang,
  Z., Dong, L., et~al. (2021c).
\newblock Swin transformer v2: Scaling up capacity and resolution.
\newblock {\em arXiv preprint arXiv:2111.09883}.

\bibitem[Liu et~al., 2021d]{liu2021swin}
Liu, Z., Lin, Y., Cao, Y., Hu, H., Wei, Y., Zhang, Z., Lin, S., and Guo, B.
  (2021d).
\newblock Swin transformer: Hierarchical vision transformer using shifted
  windows.
\newblock {\em arXiv preprint arXiv:2103.14030}.

\bibitem[Liu et~al., 2021e]{liu2021video}
Liu, Z., Ning, J., Cao, Y., Wei, Y., Zhang, Z., Lin, S., and Hu, H. (2021e).
\newblock Video swin transformer.

\bibitem[Long et~al., 2015]{long2015fully}
Long, J., Shelhamer, E., and Darrell, T. (2015).
\newblock Fully convolutional networks for semantic segmentation.
\newblock In {\em Proceedings of the IEEE conference on computer vision and
  pattern recognition}, pages 3431--3440.

\bibitem[Radford et~al., 2021]{radford2021clip}
Radford, A., Kim, J.~W., Hallacy, C., Ramesh, A., Goh, G., Agarwal, S., Sastry,
  G., Askell, A., Mishkin, P., Clark, J., et~al. (2021).
\newblock Learning transferable visual models from natural language
  supervision.
\newblock In {\em International Conference on Machine Learning}, pages
  8748--8763. PMLR.

\bibitem[Rebuffi et~al., 2017]{rebuffi2017resadapt}
Rebuffi, S.-A., Bilen, H., and Vedaldi, A. (2017).
\newblock Learning multiple visual domains with residual adapters.
\newblock {\em Advances in neural information processing systems}, 30.

\bibitem[Rebuffi et~al., 2018]{rebuffi2018multidomain}
Rebuffi, S.-A., Bilen, H., and Vedaldi, A. (2018).
\newblock Efficient parametrization of multi-domain deep neural networks.
\newblock In {\em Proceedings of the IEEE Conference on Computer Vision and
  Pattern Recognition}, pages 8119--8127.

\bibitem[Rusu et~al., 2016]{rusu2016progressive}
Rusu, A.~A., Rabinowitz, N.~C., Desjardins, G., Soyer, H., Kirkpatrick, J.,
  Kavukcuoglu, K., Pascanu, R., and Hadsell, R. (2016).
\newblock Progressive neural networks.
\newblock {\em arXiv preprint arXiv:1606.04671}.

\bibitem[Schuhmann et~al., 2021]{schuhmann2021laion}
Schuhmann, C., Vencu, R., Beaumont, R., Kaczmarczyk, R., Mullis, C., Katta, A.,
  Coombes, T., Jitsev, J., and Komatsuzaki, A. (2021).
\newblock Laion-400m: Open dataset of clip-filtered 400 million image-text
  pairs.
\newblock {\em arXiv preprint arXiv:2111.02114}.

\bibitem[Sermanet et~al., 2013]{sermanet2013overfeat}
Sermanet, P., Eigen, D., Zhang, X., Mathieu, M., Fergus, R., and LeCun, Y.
  (2013).
\newblock Overfeat: Integrated recognition, localization and detection using
  convolutional networks.
\newblock {\em arXiv preprint arXiv:1312.6229}.

\bibitem[Sharif~Razavian et~al., 2014]{sharif2014cnn}
Sharif~Razavian, A., Azizpour, H., Sullivan, J., and Carlsson, S. (2014).
\newblock Cnn features off-the-shelf: an astounding baseline for recognition.
\newblock In {\em Proceedings of the IEEE conference on computer vision and
  pattern recognition workshops}, pages 806--813.

\bibitem[Simonyan and Zisserman, 2014]{simonyan2014two}
Simonyan, K. and Zisserman, A. (2014).
\newblock Two-stream convolutional networks for action recognition in videos.
\newblock In {\em Advances in neural information processing systems}, pages
  568--576.

\bibitem[Simonyan and Zisserman, 2015]{simonyan2014vgg}
Simonyan, K. and Zisserman, A. (2015).
\newblock Very deep convolutional networks for large-scale image recognition.
\newblock In {\em International Conference on Learning Representations}.

\bibitem[Sohn et~al., 2020]{sohn2020simple_semi}
Sohn, K., Zhang, Z., Li, C.-L., Zhang, H., Lee, C.-Y., and Pfister, T. (2020).
\newblock A simple semi-supervised learning framework for object detection.
\newblock In {\em arXiv:2005.04757}.

\bibitem[Tan et~al., 2020]{tan2020efficientdet}
Tan, M., Pang, R., and Le, Q.~V. (2020).
\newblock Efficientdet: Scalable and efficient object detection.
\newblock In {\em Proceedings of the IEEE/CVF conference on computer vision and
  pattern recognition}, pages 10781--10790.

\bibitem[Vasconcelos et~al., 2022]{vasconcelos2022proper}
Vasconcelos, C., Birodkar, V., and Dumoulin, V. (2022).
\newblock Proper reuse of image classification features improves object
  detection.
\newblock {\em arXiv preprint arXiv:2204.00484}.

\bibitem[Wei et~al., 2022]{yixuan2022icar}
Wei, Y., Cao, Y., Zhang, Z., Yao, Z., Xie, Z., Hu, H., and Guo, B. (2022).
\newblock icar: Bridging image classification and image-text alignment for
  visual recognition.

\bibitem[Xiao et~al., 2018]{xiao2018upernet}
Xiao, T., Liu, Y., Zhou, B., Jiang, Y., and Sun, J. (2018).
\newblock Unified perceptual parsing for scene understanding.
\newblock In {\em Proceedings of the European Conference on Computer Vision
  (ECCV)}, pages 418--434.

\bibitem[Xie et~al., 2021]{xie2021simmim}
Xie, Z., Zhang, Z., Cao, Y., Lin, Y., Bao, J., Yao, Z., Dai, Q., and Hu, H.
  (2021).
\newblock Simmim: A simple framework for masked image modeling.
\newblock {\em arXiv preprint arXiv:2111.09886}.

\bibitem[Xu et~al., 2021]{xu2021softteacher}
Xu, M., Zhang, Z., Hu, H., Wang, J., Wang, L., Wei, F., Bai, X., and Liu, Z.
  (2021).
\newblock End-to-end semi-supervised object detection with soft teacher.
\newblock {\em Proceedings of the IEEE/CVF International Conference on Computer
  Vision (ICCV)}.

\bibitem[Yao et~al., 2021]{yao2021cpt}
Yao, Y., Zhang, A., Zhang, Z., Liu, Z., Chua, T.-S., and Sun, M. (2021).
\newblock Cpt: Colorful prompt tuning for pre-trained vision-language models.
\newblock {\em arXiv preprint arXiv:2109.11797}.

\bibitem[Zhang et~al., 2020]{zhang2020side}
Zhang, J.~O., Sax, A., Zamir, A., Guibas, L., and Malik, J. (2020).
\newblock Side-tuning: a baseline for network adaptation via additive side
  networks.
\newblock In {\em European Conference on Computer Vision}, pages 698--714.
  Springer.

\bibitem[Zhou et~al., 2018]{zhou2018ade}
Zhou, B., Zhao, H., Puig, X., Xiao, T., Fidler, S., Barriuso, A., and Torralba,
  A. (2018).
\newblock Semantic understanding of scenes through the ade20k dataset.
\newblock {\em International Journal on Computer Vision}.

\bibitem[Zhou et~al., 2021]{zhou2021vprompt}
Zhou, K., Yang, J., Loy, C.~C., and Liu, Z. (2021).
\newblock Learning to prompt for vision-language models.
\newblock {\em arXiv preprint arXiv:2109.01134}.

\end{thebibliography}

\newpage
\appendix

\section{Memory and Speed Analysis}
We carefully analyzed the training speed and memory usage of the frozen backbone with additional tunable parameters and full finetuning on COCO object detection, ADE20K semantic segmentation and K400 video action recognition using various sized models from SwinV2-T to SwinV2-G. As shown in Table \ref{tab:memory} and \ref{tab:speed}, even with more trainable parameters in the head network, training with a frozen backbone can significantly improve speed and reduce memory cost, especially for large-scale models. Moreover, freezing the backbone reduces the memory consumption of a billion-level model to less than 32G, which makes it possible to run on regular GPUs, thus helping institutions with limited resources to take advantage of such large models.
\begin{table}[htb]
        \centering
        \small
        \addtolength{\tabcolsep}{-2.5pt}
        \begin{tabular}{c|ccc|ccc|ccc}
            \Xhline{1.0pt}
            \multirow{3}{*}{Base Network} & \multicolumn{3}{c|}{COCO} & \multicolumn{3}{c|}{ADE20K} & \multicolumn{3}{c}{Kinetics-400}\\
            & Batch & Frozen & Full ft. & Batch & Frozen & Full ft. & Batch & Frozen & Full ft. \\
            & Size & $5\times$ BiFPN & FPN & Size & $6\times$ Pix. Dec. & $1\times$ Pix. Dec. & Size & $4\times$  Blocks & Linear \\
            \hline
            SwinV2-T & 16 & 9.74G & 9.78G & 16 & 4.24G & 5.04G & 64 & 10.43G & 15.87G \\
            SwinV2-B & 16 & 10.38G & 17.29G & 16 & 4.42G & 7.84G & 64 & 13.46G & 31.47G \\
            SwinV2-L & 16 & 10.56G & 25.81G & 16 & 4.84G & 11.77G & 32 & 12.80G & 27.05G \\
            SwinV2-G & 8 & 31.05G & >80G & 8 & 23.84G & 78.77G & 16 & 30.54G & >80G \\
          \Xhline{1.0pt}  
       \end{tabular}
     \caption{Memory usage for models of different sizes.}
     \label{tab:memory}
\end{table}
\begin{table}[htb]
        \centering
        \small
        \addtolength{\tabcolsep}{-2.5pt}
        \begin{tabular}{c|ccc|ccc|ccc}
            \Xhline{1.0pt}
            \multirow{3}{*}{Base Network} & \multicolumn{3}{c|}{COCO} & \multicolumn{3}{c|}{ADE20K} & \multicolumn{3}{c}{Kinetics-400}\\
            & Batch & Frozen & Full ft. & Batch & Frozen & Full ft. & Batch & Frozen & Full ft. \\
            & Size & $5\times$ BiFPN & FPN & Size & $6\times$ Pix. Dec. & $1\times$ Pix. Dec. & Size & $4\times$  Blocks & Linear \\
            \hline
            SwinV2-T & 16 & 0.44s & 0.46s & 16 & 0.28s & 0.34s & 64 & 0.31s & 0.51s \\
            SwinV2-B & 16 & 0.50s & 0.65s & 16 & 0.31s & 0.41s & 64 & 0.52s & 0.96s \\
            SwinV2-L & 16 & 0.57s & 0.84s & 16 & 0.34s & 0.48s & 32 & 0.45s & 0.78s \\
            SwinV2-G & 8 & 1.20s & - & 8 & 1.23s & 3.08s & 16 & 1.14s & - \\
          \Xhline{1.0pt}  
       \end{tabular}
     \caption{Speed of each iteration for models of different sizes.}
     \label{tab:speed}
\end{table}

\section{ADE20K with UPerNet vs. Mask2former}

For ADE20K semantic segmentation, we further adopt UPerNet~\cite{xiao2018upernet}, another widely used framework, to conduct comparisons on different pretrained models.
We first evaluate four different pretrained Swin Transformers~\cite{liu2021swin} with UPerNet, including supervised pretraining on ImageNet-1K (SUP-1K), supervised pretraining on ImageNet-22K (SUP-22K), contrastive learning of EsViT\cite{esvit} on ImageNet-1K (EsViT-1K), and masked image modeling of SimMIM~\cite{xie2021simmim} on ImageNet-1K (SimMIM-1K). 
Results are shown in Table \ref{tab:upernet_pretrain_task}. Similar to the results of Mask2former, we can find that the SUP-22K model works the best in both frozen and finetuning settings. The SUP-1K and the EsViT-1K models have competitive results. The SimMIM-1K model achieves performance similar to the SUP-1K model in the finetuning setting but lags behind other models in the frozen setting.

\begin{table}[htb]
        \centering
        \addtolength{\tabcolsep}{-1.0pt}
        \begin{tabular}{c|cc|cc}
            \Xhline{1.0pt}
            \multirow{2}{*}{Approach} & \multicolumn{2}{c|}{Frozen} & \multicolumn{2}{c}{Full ft.} \\
            & mIoU & mIoU (ms+flip) & mIoU & mIoU (ms+flip) \\
            \hline
            SUP-1K & 43.9 & 45.5 & 49.3 & 50.2 \\
            SUP-22K & 48.8 & 50.0 & 51.3 & 52.2 \\
            EsViT-1K & 41.8 & 43.4 & 48.8 & 49.7 \\
            SimMIM-1K & 26.0 & 27.6 & 48.6 & 49.3 \\
          \Xhline{1.0pt}  
       \end{tabular}
     \caption{Comparisons of different pretraining tasks on frozen and full finetuning settings with SwinV2-B as the base network. An UPerNet framework is adopted. Results of mIoU for ADE20K semantic segmentation are reported. SUP denotes supervised classification as pretraining. \emph{ms+flip} denotes multi-scale testing with horizontal flip augmentation.}
     \label{tab:upernet_pretrain_task}
\end{table}

Even though the SUP-22K model preforms best among different pretrained models, there is a gap of -2.5 mIoU between the frozen setting and full finetuning. We thus add more parameters in the head network and see if this could close the gap. As UPerNet has an FPN-like head network, we add parameters by replacing FPN with BiFPN. As shown in Table \ref{tab:upernet_param_at_head_ade}, with a 5-layer BiFPN, the performance gap between the frozen setting and the full finetuning is reduced to -0.2 mIoU.

\begin{table}[h]
        \centering
        \addtolength{\tabcolsep}{3.0pt}
        \begin{tabular}{c|cc|cc}
            \Xhline{1.0pt}
            \multirow{2}{*}{Head Network} & \multicolumn{2}{c|}{Frozen} & \multicolumn{2}{c}{Full ft.} \\
             & mIoU & mIoU (ms+flip) & mIoU & mIoU (ms+flip) \\
            \hline
            FPN & 48.8 & 50.0 & 51.3 & 52.2 \\
            $1\times$ BiFPN Layer & 49.3 & 50.5 & 51.4 & 52.2  \\
            $2\times$ BiFPN Layer & 49.8 & 50.8 & 51.4 & 52.3 \\
            $3\times$ BiFPN Layer & 50.8 & 51.4 & 51.5 & 52.2 \\
            $4\times$ BiFPN Layer & 50.7 & 51.9 & 51.6 & 52.5 \\
            $5\times$ BiFPN Layer & 51.3 & 52.6 & 51.5 & 52.4 \\
          \Xhline{1.0pt}  
       \end{tabular}
       \caption{Comparisons with different head networks on ADE20K semantic segmentation: FPN or BiFPN with different layers, using UperNet as the segmentation framework and SwinV2-B with SUP-22K training as the base network. \emph{ms+flip} denotes multi-scale testing with horizontal flip augmentation.}
    \label{tab:upernet_param_at_head_ade}
\end{table}

\section{CKA Analysis on Mask2Former}
\begin{figure}[h]
    \centering
    \includegraphics[width=.8\linewidth]{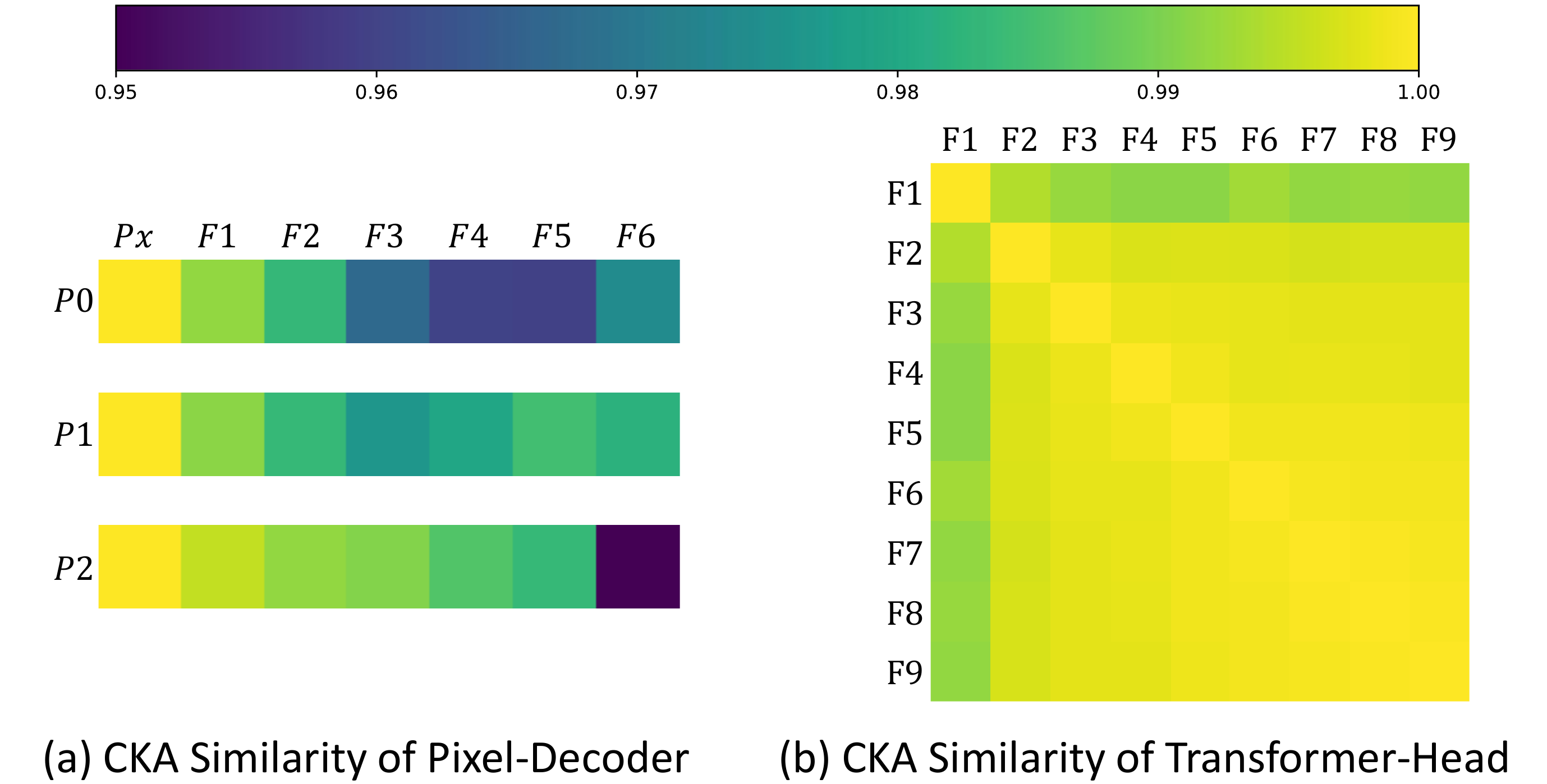}
    \caption{CKA heatmap on the features across different layers in the pixel decoder (a) and Transformer decoder (b) in Mask2former on ADE20K semantic segmentation.}
    \label{fig:ade_cka}
\end{figure}

To further understand the behavior of Mask2Former on ADE20K semantic segmentation, we conduct a similar Centered Kernel Alignment (CKA)~\cite{cka} analysis of the feature similarity across different layers in the pixel decoders and Transformer decoders as done for object detection on COCO. As show in Figure~\ref{fig:ade_cka} (a), for each stage (with different resolutions of feature map) of the pixel decoder, we plot the CKA similarity between the input features ($Px$ of stage $x$) and the output of each block ($Fx$ of block $x$). For the Transformer decoder, as shown in Figure~\ref{fig:ade_cka} (b), we plot the CKA similarity between the output features ($Fx$ of head $x$) of each Transformer decoder. 
From this figure, we can observe that the features across heads in the Transformer decoder are almost the same. But in the pixel decoder, features across different blocks are somewhat different (with lower CKA similarity). In other words, adding parameters in the pixel decoder provides more useful capacity for transformation, but adding parameters in the Transformer decoder provides much less useful computation. This well matches the previous observation on COCO with BiFPN and cascade head (shown in Figure 3 of the main paper).

\section{Detailed Settings}

\paragraph{Object Detection and Instance Segmentation on COCO} 

Object detection and instance segmentation experiments are conducted on COCO 2017~\cite{lin2014coco}. We adopt a Mask R-CNN framework and the large-scale jittering augmentation~\cite{simple_copy_paste}. Detailed configurations are listed in Table~\ref{table:setting-coco}.

\begin{table}[h]\small
    \centering
    \begin{tabular}{c|cc|cc|cc}
    \toprule
    \multirow{2}{*}{\textbf{Hyperparameters}} & \multicolumn{2}{c|}{\textbf{1\%}} & \multicolumn{2}{c|}{\textbf{10\%}} & \multicolumn{2}{c}{\textbf{100\%}} \\
    & \textbf{Frozen} & \textbf{Full Ft.} & \textbf{Frozen} & \textbf{Full Ft.} & \textbf{Frozen} & \textbf{Full Ft.}\\
    \hline
    Detector & \multicolumn{2}{c|}{Mask RCNN} & \multicolumn{2}{c|}{Mask RCNN} & \multicolumn{2}{c}{Mask RCNN} \\
    Training input size & \multicolumn{2}{c|}{(1024, 1024)} & \multicolumn{2}{c|}{(1024, 1024)} & \multicolumn{2}{c}{(1024, 1024)} \\
    Training scale ratio & \multicolumn{2}{c|}{(0.1, 2.0)} & \multicolumn{2}{c|}{(0.1, 2.0)} & \multicolumn{2}{c}{(0.1, 2.0)} \\
    Rand horizontal flip & \multicolumn{2}{c|}{0.5} & \multicolumn{2}{c|}{0.5} & \multicolumn{2}{c}{0.5} \\
    Test input size & \multicolumn{2}{c|}{(1333, 800)} & \multicolumn{2}{c|}{(1333, 800)} & \multicolumn{2}{c}{(1333, 800)} \\
    \hline
    Training epochs & \multicolumn{2}{c|}{12} & \multicolumn{2}{c|}{12} & \multicolumn{2}{c}{72} \\
    Warm-up iterations & \multicolumn{2}{c|}{500} & \multicolumn{2}{c|}{500} & \multicolumn{2}{c}{500} \\
    Batch size & \multicolumn{2}{c|}{16} & \multicolumn{2}{c|}{16} & \multicolumn{2}{c}{16} \\
    Layer decay & \xmark & 0.95 & \xmark & 0.95 & \xmark & 0.95 \\
    Base learning rate & 1e-3 & 1e-5 & 3e-4 & 3e-5 & 3e-4 & 3e-5 \\
    Weight decay & 0.5 & 0.1 & 0.1 & 0.05 & 0.05 & 0.05 \\
    Optimizer & \multicolumn{2}{c|}{AdamW} & \multicolumn{2}{c|}{AdamW} & \multicolumn{2}{c}{AdamW} \\
    Adam $\beta$ & \multicolumn{2}{c|}{(0.9, 0.999)} & \multicolumn{2}{c|}{(0.9, 0.999)} & \multicolumn{2}{c}{(0.9, 0.999)} \\
    Learning rate scheduler & \multicolumn{2}{c|}{Multi-Step}  & \multicolumn{2}{c|}{Multi-Step}  & \multicolumn{2}{c}{Multi-Step}  \\
    Step $\gamma$ & \multicolumn{2}{c|}{0.1} & \multicolumn{2}{c|}{0.1} & \multicolumn{2}{c}{0.1} \\
    Step epochs & \multicolumn{2}{c|}{(8, 11)} & \multicolumn{2}{c|}{(8, 11)} & \multicolumn{2}{c}{(63, 69)} \\
    Stochastic depth & \multicolumn{2}{c|}{0.3} & \multicolumn{2}{c|}{0.3} & \multicolumn{2}{c}{0.3} \\
    \bottomrule
    \end{tabular}
    \vspace{0.5em}
    \caption{Hyperparameters for the frozen setting and the full finetuning on COCO object detection dataset with 1\%, 10\% and all training data.}
    \label{table:setting-coco}
\end{table}

\paragraph{Semantic Segmentation on ADE20K}

For the semantic segmentation task, we adopt widely-used ADE20K~\cite{zhou2018ade} as the benchmark. Experiments are conducted with both Mask2former~\cite{mask2former} and UPerNet~\cite{xiao2018upernet}. All experiments follow the settings listed in Table~\ref{table:setting-ade} except for those with the SimMIM-1K pretrained model. For SimMIM-1K with Mask2former, we use a learning rate of 3e-4 for full finetuning. For SimMIM-1K with UPerNet, we use a learning rate of 2e-4 and a weight decay of 0.05 for full finetuning.  

\begin{table}[h]\small
    \centering
    \begin{tabular}{c|cc|cc}
    \toprule
    \multirow{2}{*}{\textbf{Hyperparameters}} & \multicolumn{2}{c|}{\textbf{Mask2former}} & \multicolumn{2}{c}{\textbf{UPerNet}} \\
    & \textbf{Frozen} & \textbf{Full Ft.} & \textbf{Frozen} & \textbf{Full Ft.} \\
    \hline
    Training input size & \multicolumn{2}{c|}{(512, 512)} & \multicolumn{2}{c}{(512, 512)} \\
    Training scale ratio & \multicolumn{2}{c|}{(0.5, 2.0)} & \multicolumn{2}{c}{(0.5, 2.0)} \\
    Rand horizontal flip & \multicolumn{2}{c|}{0.5} & \multicolumn{2}{c}{0.5} \\
    PhotoMetricDistortion & \multicolumn{2}{c|}{\cmark} & \multicolumn{2}{c}{\cmark} \\
    Test input size & \multicolumn{2}{c|}{(2048, 512)} &  \multicolumn{2}{c}{(2048, 512)} \\
    \hline
    Training iterations & \multicolumn{2}{c|}{160,000} & \multicolumn{2}{c}{160,000} \\
    Warm-up iterations & \multicolumn{2}{c|}{0} & \multicolumn{2}{c}{1,500} \\
    Batch size & \multicolumn{2}{c|}{16} & \multicolumn{2}{c}{16} \\
    Backbone lr ratio & \xmark & 0.1 & \xmark & \xmark  \\
    Layer decay & \xmark & \xmark & \xmark & 0.95 \\
    Base learning rate & 3e-4 & 1e-4 & 2e-3 & 2e-5 \\
    Weight decay & 0.02 & 0.05 & 0.05 & 0.01 \\
    Optimizer & \multicolumn{2}{c|}{AdamW} & \multicolumn{2}{c}{AdamW} \\
    Adam $\beta$ & \multicolumn{2}{c|}{(0.9, 0.999)} & \multicolumn{2}{c}{(0.9, 0.999)} \\
    Learning rate scheduler & \multicolumn{2}{c|}{Linear} & \multicolumn{2}{c}{Linear} \\
    Stochastic depth & \multicolumn{2}{c|}{0.3} & \multicolumn{2}{c}{0.3} \\
    \bottomrule
    \end{tabular}
    \vspace{0.5em}
    \caption{Hyperparameters for the frozen setting and full finetuning on ADE20K semantic segmentation.}
    \label{table:setting-ade}
\end{table}

\paragraph{Video Action Recognition on Kinetics-400}

Video action recognition experiments are evaluated on the Kinetics-400 dataset. We follow Video Swin Transformer~\cite{liu2021video} for most of the settings. Detailed hyperparameters are shown in Table~\ref{table:setting-video}.

\begin{table}[h]\small
    \centering
    \begin{tabular}{c|cc}
    \toprule
    \textbf{Hyperparameters} & \textbf{Froze} & \textbf{Full Ft.} \\
    \hline
    Training Input size & \multicolumn{2}{c}{(16, 224, 224)} \\
    Patch size & \multicolumn{2}{c}{(1, 4, 4)} \\
    Rand horizontal flip & \multicolumn{2}{c}{0.5} \\
    Rand resized crop & \multicolumn{2}{c}{\cmark} \\
    Training scale ratio & \multicolumn{2}{c}{(0.5, 2.0)} \\
    Test view & \multicolumn{2}{c}{4$\times$3} \\
    \hline
    Training epochs & \multicolumn{2}{c}{30} \\
    Warm-up epochs & \multicolumn{2}{c}{2.5} \\
    Batch size & \multicolumn{2}{c}{64} \\
    Base learning rate & \multicolumn{2}{c}{3e-4} \\
    Weight decay & \multicolumn{2}{c}{0.05} \\
    Optimizer & \multicolumn{2}{c}{AdamW} \\
    Adam $\beta$ & \multicolumn{2}{c}{(0.9, 0.999)} \\
    Learning rate scheduler & \multicolumn{2}{c}{Cosine} \\
    Stochastic depth & \multicolumn{2}{c}{0.2} \\
    \bottomrule
    \end{tabular}
    \vspace{0.5em}
    \caption{Hyperparameters for the frozen setting and full finetuning on Kinetics-400 video action recognition.}
    \label{table:setting-video}
\end{table}

\paragraph{Swin-G Setting}

To explore the performance upper bound under the frozen setting, we use a strong training setting for the Swin-G model.
For COCO object detection, we adopt a framework of HTC~\cite{chen2019htc,liu2021swin} and a large scale jittering augmentation with an input size of (1024, 1024). The window size is set as 32, the learning rate is 6e-4, and the batch size is 32. A weight decay of 0.05 and a stochastic depth rate of 0.3 are used.
For ADE20K semantic segmentation, we set the image input size as (640, 640) and the window size as 40. The learning rate is 3e-4 and the batch size is 16. A weight decay of 0.02 and a stochastic depth rate of 0.3 are used.
For Kinetics-400 video action recognition, we use a window size of 16, a learning rate of 2e-4, a stochastic depth of 0.1 and a batch size of 128. 

\end{document}